\documentclass[sigconf]{acmart}

\renewcommand\footnotetextcopyrightpermission[1]{} % no copyright
\acmConference{}{}{} % clear conference
\acmBooktitle{} % clear booktitle
\acmYear{} % clear year
\acmDOI{} % clear DOI
\acmISBN{} % clear ISBN

\AtBeginDocument{%
  }

\usepackage{xcolor}

\usepackage[linesnumbered,ruled,vlined]{algorithm2e} % put this in your preamble

\usepackage{subcaption}
\usepackage{enumitem}
\usepackage[font=small]{caption}

\begin{document}

%%
%% The "title" command has an optional parameter,
%% allowing the author to define a "short title" to be used in page headers.
\title{FusionAdapter for Few-Shot Relation Learning in Multimodal Knowledge Graphs}
%
% The "author" command and its associated commands are used to define
% the authors and their affiliations.
% Of note is the shared affiliation of the first two authors, and the
% "authornote" and "authornotemark" commands
% used to denote shared contribution to the research.
\author{Ran Liu}
\email{ran.liu.2020@phdcs.smu.edu.sg}
% \orcid{1234-5678-9012}
% \author{G.K.M. Tobin}
% \authornotemark[1]
% \email{webmaster@marysville-ohio.com}
\affiliation{%
  \institution{Singapore Management University}
  % \city{Dublin}
  % \state{Ohio}
  \country{Singapore}
}

\author{Yuan Fang}
\email{yfang@@smu.edu.sg}
% \orcid{1234-5678-9012}
% \author{G.K.M. Tobin}
% \authornotemark[1]
% \email{webmaster@marysville-ohio.com}
\affiliation{%
  \institution{Singapore Management University}
  % \city{Dublin}
  % \state{Ohio}
  \country{Singapore}
}

\author{Xiaoli Li}
\email{xiaoli_li@sutd.edu.sg}
% \orcid{1234-5678-9012}
% \author{G.K.M. Tobin}
% \authornotemark[1]
% \email{webmaster@marysville-ohio.com}
\affiliation{%
  \institution{Singapore University of Technology and Design}
  % \city{Dublin}
  % \state{Ohio}
  \country{Singapore}
}

% %%
%% By default, the full list of authors will be used in the page
%% headers. Often, this list is too long, and will overlap
%% other information printed in the page headers. This command allows
%% the author to define a more concise list
%% of authors' names for this purpose.
% \renewcommand{\shortauthors}{Trovato et al.}
\newcommand{\model}{FusionAdapter}
\newcommand{\stitle}[1]{\vspace{1.5mm}\noindent\textbf{#1}\hspace{0.5em}}

%%
%% The abstract is a short summary of the work to be presented in the
%% article.
\begin{abstract}
 Multimodal Knowledge Graphs (MMKGs) incorporate various modalities, including text and images, to enhance entity and relation representations. Notably, different modalities for the same entity often present complementary and diverse information. 
    However, existing MMKG methods primarily align modalities into a shared space, which tends to overlook the distinct contributions of specific modalities, limiting their performance particularly in low-resource settings. To address this challenge, we propose \model for the learning of few-shot relationships (FSRL) in MMKG.
    \model\ introduces (1) an adapter module that enables efficient adaptation of each modality to unseen relations and (2) a fusion strategy that integrates multimodal entity representations while preserving diverse modality-specific characteristics. By effectively adapting and fusing information from diverse modalities, \model\ improves generalization to novel relations with minimal supervision. Extensive experiments on two benchmark MMKG datasets demonstrate that \model\ achieves superior performance over state-of-the-art methods. (Code available in anonymous link\footnote{\scriptsize \url{https://anonymous.4open.science/r/FusionAdapter-EFCA}}.)
\end{abstract}

%%
%% The code below is generated by the tool at http://dl.acm.org/ccs.cfm.
%% Please copy and paste the code instead of the example below.
%%
% \begin{CCSXML}
% <ccs2012>
%    <concept>
%        <concept_id>10010147.10010178.10010187.10010188</concept_id>
%        <concept_desc>Computing methodologies~Semantic networks</concept_desc>
%        <concept_significance>500</concept_significance>
%        </concept>
%    <concept>
%        <concept_id>10002951.10003260</concept_id>
%        <concept_desc>Information systems~World Wide Web</concept_desc>
%        <concept_significance>500</concept_significance>
%        </concept>
%  </ccs2012>
% \end{CCSXML}

% \ccsdesc[500]{Computing methodologies~Semantic networks}
% \ccsdesc[500]{Information systems~World Wide Web}

%%
%% Keywords. The author(s) should pick words that accurately describe
%% the work being presented. Separate the keywords with commas.
\keywords{Knowledge graphs, Few-shot learning, Meta-learning, Multimodal fusion, Parameter-efficient learning}

%\received{20 February 2007}
%\received[revised]{12 March 2009}
%\received[accepted]{5 June 2009}

%%
%% This command processes the author and affiliation and title
%% information and builds the first part of the formatted document.
\maketitle

\section{Introduction}

% Knowledge graphs (KGs) are widely adopted on the modern Web, offering opportunities to organize and interlink information at Web scale \cite{hogan2021knowledge,dong2014knowledge}. KGs play a pivotal role in powering key Web applications such as search engines and recommendation systems, transforming how users discover and consume information on the Web with improved relevance. By providing a unified, machine-readable knowledge representation, KGs establish a principled foundation for information retrieval, large-scale data integration, and AI-driven personalization on the Web.

Knowledge graphs (KGs) are widely adopted on the modern Web, offering opportunities to organize and interlink information at scale \cite{hogan2021knowledge,dong2014knowledge}. KGs play a pivotal role in powering applications such as search engines and recommendation systems, transforming how users discover and consume information with improved relevance. By providing a unified, machine-readable knowledge representation, KGs establish a principled foundation for information retrieval, large-scale data integration, and AI-driven personalization.

Despite their success, conventional KGs \cite{bollacker2008freebase,suchanek2007yago,yu2023web} represent real-world facts as symbolic triples of the form (head entity, relation, tail entity). This formulation is effective for capturing facts, but it remains limited in its ability to express the diverse signals and complex characteristics of entities that arise from multimodal Web data. Recently, multimodal knowledge graphs (MMKGs) \cite{liang2024survey,kannan2020multimodal} have emerged, leveraging various data modalities to enrich entity and relation representations. For example, integrating textual descriptions and images provides complementary information to triples, enhancing a wide range of applications such as question answering \cite{speiser2022visual}, recommendation systems \cite{liu2023megcf}, and object recognition \cite{tan2019lxmert}. 

\begin{figure}[t]
  \centering
  \includegraphics[width=\linewidth]{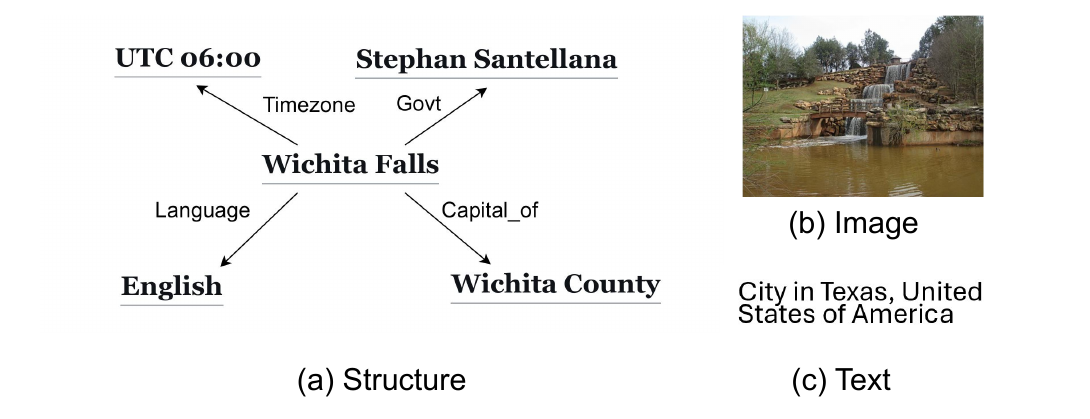}
  \caption{Diverse information across different modalities for the entity ``Wichita Falls'' in the FB-IMG dataset.}
  \label{fig:pilot}
\end{figure}

\stitle{Limitations of Prior Work.} Most MMKG approaches focus on aligning different modalities into a shared representation space \cite{chen2024knowledge, chen2020uniter, speiser2022visual, liu2023megcf, tan2019lxmert, cao2022otkge}. However, this often leads to over-homogenization and the loss of valuable modality-specific information.
 
Figure~\ref{fig:pilot} illustrates the complementary nature of multiple modalities in representing the entity “Wichita Falls.” Each modality captures a unique aspect of the entity, yet often provides partial information. The structural modality conveys relational context but omits explicit visual or geographic details, which are instead supplied by the image and textual modalities. However, prior approaches that rely on rigid alignment \cite{chen2024knowledge,chen2020uniter} struggle to fully exploit these modality-specific characteristics. Such misalignment can introduce noise and distortions, ultimately degrading performance \cite{hama2023multi}. This highlights the open challenge of integrating multimodal data in a way that preserves the individuality of each modality while enabling effective fusion.

Moreover, existing approaches focus on supervised settings, while few-shot relation learning (FSRL) \cite{xiong2018one} remains largely unexplored in MMKGs. Given an incomplete KG, FSRL aims to infer unseen relations from a few examples, which is a common real-world scenario. However, previous FSRL efforts \cite{xiong2018one,chen2019meta} are developed for unimodal KGs only. Meanwhile, with few examples available, preserving complementary information from different modalities is crucial for effective relation prediction.

\stitle{Problem and insights.}
In this work, we study FSRL in MMKGs, for which we must address two challenges. The first challenge (\textbf{C1}) is how to perform multimodal fusion while preserving the distinct characteristics of each modality. The second challenge (\textbf{C2}) is achieving this in a parameter-efficient manner for few-shot settings, as excessive parameters \cite{liang2022modular} can hinder generalization to unseen relations.

To address the above challenges, we introduce \model, a multimodal fusion framework for FSRL. For \textbf{C1}, we propose a diversity loss that prevents rigid modality alignment, ensuring each modality retains its distinct information. It enables the model to extract complementary features rather than collapsing them into an averaged representation. Additionally, it improves robustness by reducing noise from modality misalignment, leading to more accurate and reliable predictions. For \textbf{C2}, we propose a lightweight adapter module for parameter-efficient multimodal fusion. During downstream testing, only the adapter's parameters are updated, improving generalization to unseen relations with few examples.

\stitle{Contributions.}
In summary, we make the following contributions in this work. 
\begin{itemize}[leftmargin=*]
    \item We observe that different modalities for the same entity often provide complementary and distinct information. Thus, we introduce a novel multimodal fusion strategy with a diversity loss to preserve the modality-specific characteristics.
    \item We propose a lightweight adapter module to fuse multimodal information in a parameter-efficient manner, enabling generalization to unseen relations with minimal examples.
    \item We conduct extensive experiments on two MMKG benchmarks, and demonstrate the superiority of our proposed \model.
\end{itemize}

\section{Related Work}
This section outlines key research for multimodal knowledge graphs and few-shot relation learning.

\stitle{MMKG aproaches.}
MMKGs extend traditional KGs by incorporating additional modalities such as text, images, and audio to enrich entity and relation representations \cite{liang2024survey}. There are four main approaches. % These models can generally be categorized into three main approaches:
(1) Translation-based methods: Inspired by TransE \cite{bordes2013translating}, some methods \cite{wang2019multimodal, xie2016image,zhang2024unleashing} integrate multimodal features into additive models to enhance entity and relation embeddings. However, they simply map all modalities to a common space and fail to retain the unique characteristics of each modality.
(2) Semantic matching-based methods: These models extend uni-modal approaches by incorporating cross-modal interactions using bilinear or tensor-based transformations \cite{garcia2017kblrn,cao2022otkge,chen2022hybrid}. They capture richer relationships between modalities but tend to introduce high computational costs and parameter inefficiency.
(3) Graph-based models: Graph neural networks (GNNs) have been adapted to MMKGs to leverage both structural information and modality-specific features. However, GNN-based methods can struggle to effectively integrate heterogeneous multi-modal data without specialized designs.\cite{lee2024multimodal}. 
(4) Retrieval-augmented methods: NativE \cite{zhang2024native} introduces a retrieval-enhanced framework that leverages external multi-modal information and aligns it to target triples. (4) Retrieval-augmented methods: NativE \cite{zhang2024native} introduces a retrieval-enhanced framework that leverages external multi-modal information and aligns it to target triples.

Despite leveraging multimodal information, these methods described above depend on supervised learning and require substantial amounts of labeled data. Moreover, existing MMKG techniques often assume strict modality alignment, which limits their ability to fully exploit the diverse and complementary information available across different modalities.

\stitle{Few-shot relation learning.}
The mainstream technique for FSRL is meta-learning \cite{hospedales2021meta}, which consists of two subcategories. 
(1) Metric-based models: These models compute a similarity score between support and query sets to learn a matching metric for classification. GMatching \cite{xiong2018one} is a pioneering one-shot relation learning model, employing a one-hop neighbor encoder and a matching network. Subsequent studies further introduce an attention mechanism to aggregate information from multiple reference triples \cite{zhang2020few}, a relation-specific adaptive neighbor encoder to better capture one-hop contextual information \cite{sheng2020adaptive}, and a multi-level matching and aggregation mechanism \cite{wei2024multi}.

(2) Optimization-based models: These models aim to learn an initial meta-prior that can be quickly adapted to unseen relations with minimal supervision. MetaR \cite{chen2019meta} is a classic framework that transfers relation-specific meta-knowledge from a support set to query examples. Subsequent works refine MetaR by incorporating a gated attentive neighbor aggregator to capture the most valuable contexts \cite{niu2021relational}, considering triple-level contextual information to enhance adaptation \cite{wu2023hierarchical}, employing an adapter mechanism to address distribution shift across relations \cite{ran-etal-2024-context}, and modeling inter-entity interactions to better capture complex contextual dependencies \cite{li2022learning}.

However, these methods operate in a unimodal setting, relying solely on structured triples from KGs. Additional modalities, such as text and images, could provide complementary information and improve generalization for unseen relations. Thus, exploring the fusion of multimodal data while preserving distinct modality-specific information is important for FSRL on MMKGs.

Our approach \model\ differs from RelAdapter \cite{ran-etal-2024-context} and other meta-learning methods in two key aspects. (1) Motivation: RelAdapter aligns relation distributions across tasks, while \model\ targets modality fusion in few-shot MMKGs, leveraging complementary signals from different modalities. (2) Methodology: \model\ employs a contrastive diversity loss to balance shared and modality-specific features, enabling effective fusion without losing unique information. These distinctions highlight \model’s suitability for multimodal few-shot relation learning.

\section{Preliminaries}\label{sec:prelim}
In this section, we introduce the problem of multimodal FSRL and the meta-learning framework for this problem.

\stitle{Problem formulation.}
A multimodal knowledge graph (MMKG) can be represented by $\mathcal{G} = (\mathcal{V}, \mathcal{R}, \mathcal{M})$, extending traditional knowledge graphs (KGs) by incorporating multiple modalities, where $\mathcal{M} = \{\mathbf{E}_{\text{S}}, \mathbf{E}_{\text{T}}, \mathbf{E}_{\text{V}}\}$ represents structural, textual, and visual embeddings of the entities, respectively. Each triplet is represented as $(h, r, t)$ for some entities $h, t \in \mathcal{V}$ and relation $r \in \mathcal{R}$. %where $\mathcal{V}$ is the set of entities and $\mathcal{R}$ is the set of relations.

Given an unseen relation $r \notin \mathcal{R}$ with respect to $\mathcal{G}$, we define a multimodal support set as \( \mathcal{S}_r = \{ (h_i, r, t_i) \mid i = 1, \dots, K \} \), where each entity consists of three modalities from \( \mathcal{M} = \{ \mathbf{E}_{\text{S}}, \mathbf{E}_{\text{T}}, \mathbf{E}_{\text{V}} \} \). The goal is to predict the missing tail entities in a multimodal query set:
$\mathcal{Q}_r = \{(h_j, r, ?) \mid j=1,2,\dots\}$,
given the head entity $h_j \in \mathcal{V}$ and the relation $r$. The corresponding candidate set %$\mathcal{C}_{h_j, r}$
is constructed by taking all possible entities $t_j \in \mathcal{V}$ as candidates for the missing tail. Together, the support and query sets form a task: $\mathcal{T}_r = (\mathcal{S}_r, \mathcal{Q}_r)$.
Since the support set provides only $K$ instances, this represents  a $K$-shot relation learning task, where $|\mathcal{S}_r| = K$ is typically small.

\stitle{Meta-learning framework.} 
%\label{sec:backbone}
Given the success of meta-learning in few-shot problems, we adopt MetaR \cite{chen2019meta}, a classic meta-learning framework for FSRL, as our foundation.
% Owing to the success of MetaR model, in this paper, we adopt it as the backbone for $\model$. 
MetaR consists of two stages: \emph{meta-training} and \emph{meta-testing}, aiming to learn a prior $\Phi$ from the meta-training stage that can be adapted to the meta-testing stage.
On one hand, meta-training involves a set of seen relations $\mathcal{R}^\text{tr}$, and operates on their task data $\mathcal{D}^{\text{tr}}  =\{\mathcal{T}_r\mid r\in \mathcal{R}^\text{tr}\}$.
On the other hand, meta-testing involves a set of unseen relations $\mathcal{R}^\text{te}$ such that $\mathcal{R}^\text{tr} \cap \mathcal{R}^\text{te}=\emptyset$, and operates on their task data   
$\mathcal{D}^{\text{te}}  =\{\mathcal{T}_r\mid r\in \mathcal{R}^\text{te}\}$. Note that the ground-truth tail entities are provided in the query sets of the meta-training tasks $\mathcal{D}^{\text{tr}}$, whereas the objective is to predict the unknown tail entities for the query sets of the meta-testing tasks $\mathcal{D}^{\text{te}}$.
\begin{figure}[tbp]
    \hspace*{-8pt}%
    \centering    \includegraphics[width=1.11\linewidth]{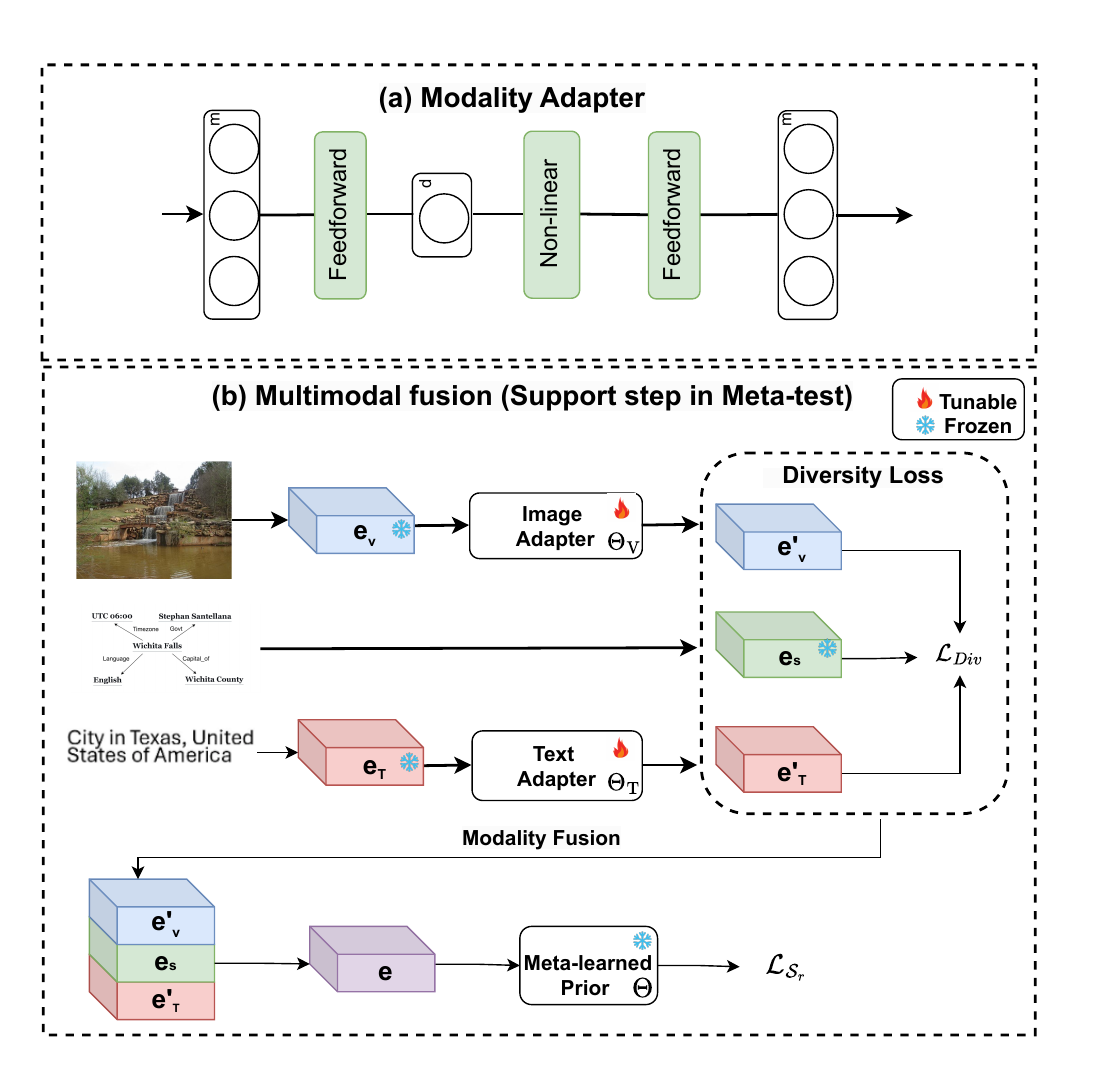}%
     \vspace{-3mm}
    \caption{Illustration of key concepts in \model.
    (a) The adapter module for modality fusion.
    (b) The fusion strategy during meta-testing.
    %hinging on an fusion-based adapter structure (a) and the fusion strategy (b) during meta-testing. 
    \emph{For brevity, we omit the meta-training stage, which is similar to meta-testing but with backpropagation of the task loss to update the model parameters.}}
    % ($\mathtt{emb}$ and $\Phi$). }
    \label{fig:framework}
\end{figure}

\section{Methodology} 

In this section, we introduce the proposed approach \model. As depicted in Fig.~\ref{fig:framework}, \model\ has two  mains components: (a) a modality adapter module, and (b) a multimodal fusion strategy. On one hand, the adapter network aims to facilitate modality fusion in a parameter-efficient manner. On the other hand, the fusion strategy integrates multimodal entity embeddings while preserving diverse modality-specific characteristics. 
%The two components work hand in hand to enhance FSRL for the novel relations in meta-testing while retaining the uniqueness of each modality.

In the rest of the section, we present each component and the overall meta-learning pipeline.
%we first introduce the adapter module, followed by the details of the meta-training and meta-testing stages. 

\subsection{Modality Adapter} \label{sec:method:adapter_architecture}
To improve the adaptation of multimodal data to unseen relations, we propose a modality adapter. The adapter transforms the textual and visual embeddings of an entity to align with the corresponding structural embeddings, facilitating multimodality fusion in the next step. Designed to be lightweight, the adapter enables parameter-efficient fusion: only its parameters are learnable for each unseen relation in meta-testing.  

%Unlike conventional projection-based fusion, which aligns all modalities into a shared space, our approach preserves modality-specific characteristics by introducing a diversity loss. This loss prevents text and image embeddings from collapsing into structural representations, ensuring that each modality retains its distinct information.

%\stitle{multimodal Adapter.}

%The adapter is responsible for processing multimodal features while maintaining diversity between structure, text, and image embeddings. Instead of mapping textual and visual embeddings directly to the structured space, the adapter applies a diversity loss that encourages the text and image embeddings to remain distinct from the structured embeddings.

Formally, for a given task $\mathcal{T}_r$, we apply a separate feed-forward layer for textual and visual embeddings. For each entity $i\in \mathcal{V}$,
\begin{align} \mathbf{e}'_{\text{T},i} &= \mathtt{Adapter}_{\text{T}}(\mathbf{e}_{\text{T},i};\Theta_{\text{T}}),\label{eq:adapter_text} \\
\mathbf{e}'_{\text{V},i} &= \mathtt{Adapter}_{\text{V}}(\mathbf{e}_{\text{V},i}; \Theta_{\text{V}}) \label{adapter},
\end{align}
where $\mathbf{e}_{\text{T},i}$ and $\mathbf{e}_{\text{V},i}$ represents the original textual and visual embeddings from  $\mathbf{E}_{\text{T}}$ and  $\mathbf{E}_{\text{V}}$, respectively,
$\mathbf{e}'_{\text{T},i}$ and $\mathbf{e}'_{ \text{V},i}$ are the adapted text and visual embeddings, respectively. 
$\Theta_{\text{T}}$ and $\Theta_{\text{V}}$ are the parameters of the text and visual adapters. Each adapter is implemented as a lightweight feed-forward network (FFN) with an efficient bottleneck structure \cite{houlsby2019parameter}, as shown in Fig.~\ref{fig:framework}(a). The output dimensions of both adapters are set equal to facilitate fusion.

% Unlike standard adapters, we do not use a residual connection to avoid forcing text and image embeddings to be too similar to their original representations.

\subsection{Multimodal Fusion}

To obtain a unified multimodal representation for each entity, we integrate the adapted textual and visual embeddings with the structural embedding, as shown in Fig.~\ref{fig:framework}(b). 
Unlike conventional projection-based fusion \cite{lee2024multimodal,cao2022otkge}, which aligns all modalities into a shared space, our approach aims to preserve modality-specific characteristics to extract complementary features rather than collapsing them into an average representation. Specifically, we propose a diversity loss that prevents excessive similarity to preserve the diverse and complementary information across modalities, while still allowing partial alignment.

%Instead of aligning all modalities to a common space, we preserve their diversity through the adapter module. 

\stitle{Contrastive diversity loss.}
To diversify the textual and visual embeddings, we introduce a contrastive diversity loss on entity embeddings. Given a relation $r$, let $\mathcal{V}_r$ denote the bag of entities from the known instances of $r$. Then, the diversity loss is 
\begin{align}
    \mathcal{L}_{\text{div}}(\mathcal{V}_r) = \frac{1}{|\mathcal{V}_r|} & \sum_{i \in \mathcal{V}_r}
    \left( [\mathtt{sim}(\mathbf{e}'_{\text{V},i}, \mathbf{e}_{\text{S},i}) - \gamma]_{+} \right.\nonumber\\
    &\left. +\ [\mathtt{sim}(\mathbf{e}'_{\text{T},i}, \mathbf{e}_{\text{S},i}) - \gamma]_{+} \right),
    \label{eqn:diversityloss}
\end{align}
where $\mathtt{sim}(\cdot, \cdot)$ represents cosine similarity, $[x]_{+}\triangleq\max(0,x)$ is a margin-based loss, and 
$\gamma$ is a hyper-parameter to control the margin. 

On one hand, the loss penalizes excessive similarity between the adapted textual or visual embeddings and the structural embeddings, encouraging their diversity and preserving modality-specific information. On the other hand, the margin $\gamma$ ensures that partial alignment can be maintained, as the penalty applies only to similarity exceeding the threshold $\gamma$. Since cosine similarity ranges from $[-1,1]$, we set $\gamma=0$ at the midpoint.

% \begin{align}
%     \mathcal{L}_{\text{div}} &= \frac{1}{|\mathcal{S}_r|} \sum_{i \in \mathcal{S}_r} 
%     \left( \max(0, \text{sim}(\mathbf{e}^{'}_{\text{V},i}, \mathbf{e}_{\text{S},i}) \right. \notag \\
%     &\quad \left. - \gamma) + \max(0, \text{sim}(\mathbf{e}^{'}_{\text{T},i}, \mathbf{e}_{\text{S},i}) - \gamma) \right)
%     \label{eqn:diversityloss}
% \end{align}

\stitle{Modality fusion.} Finally, the fused entity embeddings are given:
\begin{align}
    \mathbf{E} = \mathbf{E}_{\text{S}} + \mathbf{E}'_{\text{T}} +
     \mathbf{E}'_{\text{V}},\label{eqn:fusion}
\end{align}
where $\mathbf{E}'_{\text{T}}$ and $\mathbf{E}'_{\text{V}}$ represent the adapted textual and visual embedding matrices  obtained through the adapter modules, respectively. %, ensuring that the text modality retains its unique characteristics while still being integrated into the final representation. Similarly, $\mathbf{E}^{'}_{\text{V}}$ denotes the adapted image embedding, which maintains the distinct visual features necessary for multimodal learning. 
The structural embedding matrix, $\mathbf{E}_{\text{S}}$, remains unchanged, serving as the primary modality to which other modalities are adapted. 
%directly taken from the knowledge graph and serves as the foundation of the entity representation. By summing these three components, 
The fused embeddings integrate and align multimodal information through adapter-based transformations, while preserving modality-specific characteristics, allowing for a more informative representation.

\subsection{Overall Meta-learning Pipeline}\label{sec:method:meta-learning}  

Following MetaR \cite{chen2019meta}, our proposed \model\ consists of the meta-training and meta-testing stages.  

\stitle{Meta-training.} The meta-training stage focuses on learning the embedding matrix and prior knowledge from the seen relations. The overall objective function consists of the task loss on the query set, $\mathcal{L}_{\mathcal{Q}_r}$, and the diversity loss, $\mathcal{L}_{\text{div}}$. During meta-training, the meta-learned prior $\Theta$, the adapters' parameters $\Theta_{\text{A}}=\{\Theta_{\text{V}},\Theta_{\text{T}}\}$, and the embedding matrices $\mathcal{M} = \{ \mathbf{E}_{\text{S}}, \mathbf{E}_{\text{T}}, \mathbf{E}_{\text{V}} \}$ are optimized by minimizing the task loss along with the diversity loss, as follows.
\begin{align}
\min_{\Theta,\Theta_{\text{A}},\mathcal{M}} \mathcal{L}_{\text{meta-train}} = 
\sum_{r\in \mathcal{R}^{\text{tr}}} \mathcal{L}_{{\mathcal{Q}_r}} + \alpha  \mathcal{L}_{\text{div}}(\mathcal{V}_r),\label{eqn:meta-train}
\end{align}
where $\alpha$ is the diversity co-efficient that controls the weight of the diversity loss, balancing the modality adaptation and the preservation of modality-specific information. The query loss,  \( \mathcal{L}_{{\mathcal{Q}_r}} \), evaluates how accurately the model scores the triples in the query set $\mathcal{Q}_r$ based on a scoring function, as follows.
\begin{align}
\mathcal{L}_{\mathcal{Q}_r} = \frac{1}{|\mathcal{Q}_r|} \hspace{-0.5mm}\sum_{(h, r, t) \in \mathcal{Q}_r} \hspace{-1.5mm} [ s(h, r, t) - s(h, r, t') + \epsilon ]_+, \label{eqn:queryloss}
\end{align}
where $s(h, r, t)\triangleq \|\mathbf{e}_h + R'_{\mathcal{T}_{\text{r}}} - \mathbf{e}_t\|$ is a scoring function to evaluate the goodness of the triple $(h,r,t)$. $\mathbf{e}_h$ and $\mathbf{e}_t$ are the fused embeddings for the entities $h$ and $t$ from $\mathbf{E}$, respectively, and $R'_{\mathcal{T}_{\text{r}}}$ denotes the ``relation meta'' \cite{chen2019meta}, which is obtained from the meta-learned prior and gradient on the support set to capture the interaction between the head and tail entities for relation $r$. $t'$ is a randomly sampled tail entity to construct a negative triple $(h,r,t')$. $\epsilon \ge 0$ is a margin hyper-parameter that enforces a boundary to distinguish between positive and negative triples. %$R'_{\mathcal{T}_{\text{r}}}$ denotes the relation meta \cite{chen2019meta}, which is obtained from the meta-learned prior and gradient on the support set, to capture the interaction between the head and tail entities for each relation. %, constructed by averaging their embeddings.

%add definition of L_Qr here.

\stitle{Meta-testing.} The objective function in the meta-testing phase incorporates two similar components: the loss on the support set, $\mathcal{L}_{\mathcal{S}_r}$, and the diversity loss $\mathcal{L}_{\text{div}}$, as follows.
\begin{align}
\min_{\Theta_A}
\mathcal{L}_{\text{meta-test}} =\sum_{r\in \mathcal{R}^{\text{te}}} \mathcal{L}_{{\mathcal{S}_r}} + \alpha \mathcal{L}_{\text{div}}(\mathcal{V}_r),\label{eqn:meta-test}
\end{align}
where $\alpha$ is the same diversity coefficient as in Eq.~\eqref{eqn:meta-train}, and the support loss $\mathcal{L}_{{\mathcal{S}_r}}$ is defined in the same way as Eq.~\eqref{eqn:queryloss}, substituting the query set with the support set.
However, only the adapters' parameters $\Theta_{\text{A}}$ are updated based on the above loss, while the meta-learned prior $\Theta$ and embedding matrices $\mathcal{M}$ remain frozen, ensuring parameter efficiency in generalizing to the unseen relation $r \in \mathcal{R}^{\text{te}}$ with only a few examples in its support set. 

After updating the adapters using the support set, we employ them to predict the missing tail entity in each query triple $(h,r,?)$ from $\mathcal{Q}_r$.
Specifically, we compute a score $s(h,r,t)$ for each candidate tail $t \in \mathcal{V}$ based on the updated adapters, and rank the resulting candidate triples in ascending order of their scores.

\begin{algorithm}[tbp]
\caption{Meta-testing for \model}
\label{algo:meta-test}
\small

\KwIn{Few-shot tasks for novel relations $\mathcal{D}^{\text{te}}$, modality set $\mathcal{M}$, prior $\Theta$}

\For{each task $\mathcal{T}_r=(S_r, Q_r) \in \mathcal{D}^{\text{te}}$}{
    \While{$\Theta_{A}$ not converged}{
        Compute adapted modalities by Eqs.~\eqref{eq:adapter_text}--\eqref{adapter}\;
        Compute fused embeddings $\mathbf{E}$ by Eq.~\eqref{eqn:fusion}\;
        Minimize $\mathcal{L}_{\text{meta-test}}$ by Eq.~\eqref{eqn:meta-test}\;
        Update $\Theta_{A}$ w.r.t.~$\mathcal{L}_{\text{meta-test}}$\;
    }
    \For{each $(h,r,?) \in \mathcal{Q}_r$}{
        Rank candidates by $s(h,r,t)$\;
    }
}
\end{algorithm}

\stitle{Algorithm.} We outline our meta-testing procedure in Algorithm~\ref{algo:meta-test}. Compared to MetaR, the novel components in \model\ include fine-tuning the modality adapters for each target relation (line 3), and
the inclusion of diversity loss within $\mathcal{L}_{\text{meta-test}}$ (line 5). The overall time complexity remains unchanged and is linear to the number of shots. The overhead of the adapter module is neglible due to its parameter-efficient design (see results in Sect.~\ref{section:efficiency}).

% \stitle{Algorithm.} We outline the algorithm of our meta-testing procedure in Appendix~\ref{algo}. 

\section{Experiments}\label{sec:expt}
In this section, we conduct comprehensive experiments to evaluate our proposed \model.
% \footnote{Anonymized codes are included in Supplementary Material for review.}.
%, including performance comparison, 
% we evaluate the performance of our proposed method through comparisons against the state-of-the-art baselines on FSRL datasets, 
% followed by detailed 
%ablation study, efficiency analysis, and sensitivity analysis.

\subsection{Experiment Setup}

% \stitle{Datasets.} We utilize two benchmark datasets, namely, WN9-IMG \cite{xie2016image} and FB-IMG \cite{mousselly2018multimodal}, as summarized in Table~\ref{table:1}. For each dataset, we remove all rare relations with fewer than 20 instances. Furthermore, we use the same pre-trained embeddings as OTKGE \cite{cao2022otkge} for both datasets. Additional details for the dataset can be found in Appendix~\ref{dataset}. \lr{Due to space constraints, we have included the experimental results for the DB15K-IMG dataset, which also demonstrates favorable performance, in Appendix~\ref{DB15K}.}

\stitle{Datasets.} We utilize two benchmark datasets, namely, \textbf{WN9-IMG} \cite{xie2016image} and \textbf{FB-IMG} \cite{mousselly2018multimodal}. Additionally, we consider a filtered version of FB-IMG, named \textbf{FB-IMG (F)}, where triples in the training and validation sets are restricted to relations contained in FB15K-237 \cite{sun2019re}, in order to mitigate potential leakage caused by inverse relations. The three datasets are summarized in Table~\ref{table:1}. Popular FSRL datasets like NELL-One and Wiki-One lack multimodal versions, and aligning modalities (e.g., text and images) is non-trivial, often requiring manual effort and risking inconsistency \cite{Shang2024LAFA}. Since we focus on methodology rather than dataset construction, we adopt established MMKG datasets with pre-aligned multimodal data, enabling consistent and reliable evaluation on few-shot splits.  On each dataset, the relations are divided into three subsets: training, validation and test, as shown in Table~\ref{table:1}. In addition, we remove all rare relations with fewer than 20 instances. Furthermore, we use the same pre-trained embeddings as OTKGE \cite{cao2022otkge}.

% Additional dataset details are provided in Appendix~\ref{dataset}.

% Due to space constraints, experimental results for DB15K-IMG are included in Appendix~\ref{E}.

\begin{table}[tbp]
\caption{Statistics of datasets.}
\label{table:1}
\centering
\small
\begin{tabular}{@{}c r r r r r r r@{}}
\toprule
Dataset & Entities & Triples & \multicolumn{4}{c}{Relations} \\
\cmidrule(lr){4-7}
 &  &  & Total & Train & Valid & Test \\
\midrule
WN9-IMG             & 6,555   & 14,399  & 9   & 5   & 1   & 3 \\
FB-IMG              & 11,757  & 345,226 & 581 & 380 & 40  & 161 \\
FB-IMG (F)   & 11,422  & 202,328 & 383 & 194 & 28  & 161 \\
\bottomrule
\end{tabular}
\end{table}

\stitle{Metrics.} We employ two popular evaluation metrics, mean reciprocal rank (MRR) and hit ratio at top $N$ (Hit@$N$) to compare our approach against the baselines. Specifically, MRR reflects the absolute ranking of the first relevant item in the list and Hit@$N$ calculates the fraction of candidate lists in which the ground-truth entity falls within the first $N$ positions. We report the results averaged over five runs with different seeds, along with the standard deviation for each experiment.

\begin{table*}[t]
\centering
\caption{Performance comparison against baselines in the five-shot setting. (Best: bolded, runners-up: underlined).}
%\vspace{-2mm}
\label{table:2} 
\small
\renewcommand{\arraystretch}{1.1}
\addtolength{\tabcolsep}{2pt}
% \fontsize{9.2}{9.2}\selectfont
%
\begin{tabular}{@{}c|ccc|ccc|ccc@{}}
  \toprule
  & \multicolumn{3}{c|}{WN9-IMG} & \multicolumn{3}{c|}{FB-IMG} & \multicolumn{3}{c}{FB-IMG (F)} \\
  \cmidrule(lr){2-4} \cmidrule(lr){5-7} \cmidrule(lr){8-10}
  Models & MRR & Hit@5 & Hit@1 & MRR & Hit@5 & Hit@1 & MRR & Hit@5 & Hit@1 \\
  \midrule
TransE     & .061$\pm$.023 & .055$\pm$.021 & .032$\pm$.011 & .051$\pm$.039 & .055$\pm$.044 & .028$\pm$.019 & .044$\pm$.030 & .052$\pm$.041 & .020$\pm$.017 \\
DistMult   & .075$\pm$.016 & .055$\pm$.019 & .047$\pm$.025 & .065$\pm$.042 & .057$\pm$.026 & .037$\pm$.028 & .057$\pm$.023 & .059$\pm$.032 & .033$\pm$.024 \\
ComplEx    & .053$\pm$.018 & .047$\pm$.025 & .028$\pm$.028 & .054$\pm$.034 & .053$\pm$.041 & .031$\pm$.021 & .049$\pm$.024 & .040$\pm$.032 & .025$\pm$.022 \\
\midrule
OTKGE      & .057$\pm$.016 & .054$\pm$.022 & .037$\pm$.013 & .099$\pm$.009 & .102$\pm$.031 & .058$\pm$.015 & .085$\pm$.014 & .092$\pm$.025 & .055$\pm$.014 \\
TransAE    & .050$\pm$.024 & .047$\pm$.019 & .033$\pm$.012 & .068$\pm$.023 & .065$\pm$.028 & .036$\pm$.019 & .051$\pm$.025 & .066$\pm$.032 & .027$\pm$.011 \\
AdaMF-MAT  & .037$\pm$.036 & .043$\pm$.036 & .029$\pm$.021 & .057$\pm$.033 & .050$\pm$.039 & .032$\pm$.022 & .062$\pm$.024 & .067$\pm$.033 & .037$\pm$.025 \\
NATIVE     & .083$\pm$.015 & .061$\pm$.023 & .053$\pm$.010 & .105$\pm$.022 & .092$\pm$.033 & .062$\pm$.019 & .076$\pm$.015 & .086$\pm$.031 & .052$\pm$.024 \\
\midrule
MetaR      & .209$\pm$.020 & .475$\pm$.044 & .083$\pm$.025 & .288$\pm$.011 & .374$\pm$.016 & .205$\pm$.028 & .246$\pm$.011 & .376$\pm$.037 & .124$\pm$.035 \\
FAAN       & .234$\pm$.035 & .501$\pm$.057 & .097$\pm$.018 & .295$\pm$.071 & .384$\pm$.084 & .203$\pm$.040 & .229$\pm$.017 & .361$\pm$.037 & .129$\pm$.034 \\
HiRe       & \underline{.252}$\pm$.039 & \underline{.531}$\pm$.045 & \underline{.107}$\pm$.014 & \underline{.325}$\pm$.091 & .401$\pm$.112 & \underline{.245}$\pm$.052 & .262$\pm$.048 & \underline{.380}$\pm$.043 & .144$\pm$.034 \\
CIAN       & .241$\pm$.047 & .507$\pm$.077 & .101$\pm$.026 & .313$\pm$.069 & .397$\pm$.062 & .225$\pm$.049 & \underline{.283}$\pm$.031 & .369$\pm$.047 & \underline{.169}$\pm$.041 \\
MMSN       & .221$\pm$.041 & .483$\pm$.054 & .087$\pm$.025 & .318$\pm$.056 & \underline{.411}$\pm$.059 & .226$\pm$.043 & .264$\pm$.045 & .371$\pm$.038 & .156$\pm$.037 \\
\midrule
\model\    & \textbf{.327}$\pm$.051 & \textbf{.588}$\pm$.084 & \textbf{.116}$\pm$.039 & \textbf{.405}$\pm$.018 & \textbf{.539}$\pm$.019 & \textbf{.289}$\pm$.010 & \textbf{.325}$\pm$.022 & \textbf{.457}$\pm$.034 & \textbf{.207}$\pm$.033 \\
\bottomrule
\end{tabular}%
\end{table*}

\stitle{Baselines.} \model~is compared with a series of baselines in three major categories:

% Details of each baseline can be found in Appendix~\ref{baselines}.

(1) \textit{\textbf{Traditional supervised methods}} are fundamental knowledge graph embedding models used as baselines in link prediction tasks. We choose several state-of-the-art \emph{Traditional supervised relation learning} methods, as follows: \textbf{TransE} \cite{bordes2013translating} models relations as translations in the embedding space, performing well on simple relational patterns but struggling with symmetric and complex relations. \textbf{DistMult} \cite{yang2014embedding} extends this by using a bilinear model with a diagonal relation matrix, effectively capturing symmetric relations but failing to model asymmetry. \textbf{ComplEx} \cite{trouillon2016complex} further enhances expressiveness by introducing complex-valued embeddings, enabling it to model both symmetric and asymmetric relations while maintaining computational efficiency. They learn one model for all the relations in a supervised manner. We follow the setup in prior studies \cite{xiong2018one,wu2023hierarchical,ran-etal-2024-context,sheng2020adaptive}, which trains on the triples combined from our training splits and the support sets of our test splits. We follow the original unimodal (structure-only) setup for these models, as they lack modules for multimodal fusion.

% To utilize visual and textual modalities for a fair comparison, we project these features to the same dimension using singular value decomposition (SVD) \cite{stewart1993early,Zhao_2024} before adding them to the structured embedding. 

(2) \textit{\textbf{Multimodal supervised methods}} integrate multiple modalities, such as text, images, and structure, to enhance knowledge graph embeddings. We choose several state-of-the-art \emph{Multimodal supervised relation learning} methods, as follows: \textbf{OTKGE} \cite{cao2022otkge} leverages optimal transport to align multimodal embeddings, ensuring better cross-modal consistency. \textbf{TransAE} \cite{wang2019multimodal} extends TransE by incorporating an autoencoder framework to fuse structural and visual features, improving relational representation. \textbf{AdaMF-MAT} \cite{zhang2024unleashing} introduces an adaptive multimodal fusion mechanism combined with a multi-head attention transformer, effectively capturing complex interactions between different modalities for more robust link prediction. \textbf{NativE} \cite{zhang2024native} introduces a retrieval-enhanced framework that leverages external multi-modal information and aligns it to target triples. However, these methods may struggle with noisy or irrelevant retrieved data, which can negatively affect completion accuracy.\emph{OTKGE} \cite{cao2022otkge}, \emph{TransAE} \cite{wang2019multimodal}, \emph{AdaMF-MAT} \cite{zhang2024unleashing} and \emph{NATIVE} \cite{zhang2024native}. They learn one model for all the relations in a supervised manner on MMKGs, with specialized multimodal alignment and fusion components. We follow the same training configuration as the traditional methods.

(3)  \textit{\textbf{Traditional few-shot methods}} are designed for few-shot relation prediction tasks, where the testing relations are previously unseen in training. We choose several state-of-the-art FSRL methods, as follows: \textbf{FAAN} \cite{sheng2020adaptive} introduces an adaptive neighbor encoder for different relation tasks. \textbf{HiRe} \cite{wu2023hierarchical} brings in a hierarchical relational learning framework which considers triplet-level contextual information in contrastive learning. \textbf{MetaR} \cite{chen2019meta} utilizes a MAML-based framework, which aims to learn a good initialization for the unseen relations, followed by an optimization-based adaptation. \textbf{CIAN} \cite{li2022learning} models inter-entity interactions within a knowledge graph to better capture complex contextual dependencies between entities, thereby improving the prediction of few-shot relations by leveraging richer relational context. \textbf{MMSN} \cite{wei2024multi} employs a multi-modal Siamese network that fuses structural, textual, and visual information, enabling better generalization to unseen relations in few-shot scenarios. Similar to traditional supervised methods, we adopt their original structure-only setting, as these models do not support multimodal fusion. While we also tested SVD-based fusion for comparison, results show that it often led to degraded performance due to poor modality-fusion mechanism. Full results are provided in Table~\ref{Comparative}, using the same data splits as in Table~\ref{table:1}.

\stitle{Implementation details.}
% with embeddings for missing modalities randomly initialized. 
For traditional supervised models (TransE, DistMult and ComplEx) and multimodal supervised models (OTKGE, TransAE, AdaMF-MAT and
NATIVE) not designed for FSRL, we follow the same settings in GMatching \cite{xiong2018one} by using all triplets in the train splits, as well as the support sets from the validation/test splits to train the models. For other FSRL models (MetaR, FAAN, HiRE, CIAN and MMSN), we carry out meta-training on train/validation split and meta-test on test split. For these baselines, we use their implementations from the codebases given in Appendix~\ref{app:codebases}.

For a fair comparison, we initialize all models with the same pre-trained entity and relation embeddings, if needed. All experiments are conducted on an RTX3090 GPU server. More details, including hyperparameter settings, can be found in Appendix~\ref{Implementation_details}.

\subsection{Performance Comparison}
Table~\ref{table:2} reports the quantitative comparison of \model\ against other baselines in the five-shot setting, i.e., the size of the support set is 5. (We study the effect of the number of shots in Sect.~\ref{sec:expt:sensitivity}). 
Overall, our proposed \model\ consistently outperforms baseline approaches across both datasets.
%, highlighting the effectiveness of incorporating a multimodal adapter for few-shot relation learning (FSRL). 
Notably, \model\ surpasses the most competitive baselines by 29.7\%, 24.6\% and 14.8\% on WN9-IMG, FB-IMG and FB-IMG (F) in terms of MRR, respectively.
More specifically, we make the following key observations.

(1) Supervised methods 
%, such as TransE, DistMult, and ComplEx, 
assume that all relations encountered during testing have been seen during training. %These models rely on the ability to optimize relation embeddings based on a large number of training samples, enabling them to capture meaningful relational patterns. 
However, these methods struggle with novel relations that emerge during testing with only a few examples. While we follow  previous work \cite{xiong2018one} by incorporating the support sets from the test splits into the training data for supervised methods, the limited number of examples for the test relations significantly degrades their performance.  

(2) Multimodal supervised methods (e.g., OTKGE) are often superior to the traditional approaches (TransE, DistMult, and ComplEx), as the former employ specialized components for multimodal alignment and fusion. While we have incorporated textual and visual embeddings into traditional approaches, the fusion employs a simple SVD, which are prone to misalignment and noise across modalities. This highlights the importance of aligning and integrating different modalities. However, the multimodal methods are still supervised and unable to cope with FSRL, resulting in suboptimal performance.

(3) FSRL methods achieve much stronger performance than supervised methods, due to their ability to generalize to unseen relations with limited examples. However, their performance still trails behind \model. Specifically, compared to MetaR which shares the meta-learning architecture as ours, \model\ achieves an average improvement of 43.1\% in MRR and 29.8\% in Hit@5. This demonstrates the importance of multimodal fusion.

%(4) \model\ demonstrates strong performance with robust stability. \new{On WN9-IMG, its standard deviations range from 0.039 to 0.084, on FB-IMG from 0.010 to 0.019 and on FB-IMG (Filtered) from 0.022 to 0.034 which is comparable to or lower than those of CIAN (0.026--0.077, 0.049--0.069 and 0.031--0.047 , respectively)}. These results highlight \model’s consistent effectiveness in few-shot multimodal settings.

\begin{table}[t]
\centering
\caption{Ablation study in the five-shot setting, reporting MRR.}
%\vspace{-2mm}
\label{ablation} 
\small
%\addtolength{\tabcolsep}{-1.3mm}
% \fontsize{9.2}{9.2}\selectfont
\begin{tabular}{c|ccc} 
  \toprule
 %  &\multicolumn{2}{c}{MRR} \\   \cmidrule(lr){2-3} 
   &  WN9-IMG  &  FB-IMG &  FB-IMG (F) \\  \midrule
%    MetaR &  .212$\pm$.019 &  .282$\pm$.035  \\\midrule
w/o $\mathcal{L}_{\text{div}}$ &   .275$\pm$.037 &    .375$\pm$.023  & .307$\pm$.028  \\
 w/o adapters &   .235$\pm$.042 &    .316$\pm$.014    &    .255$\pm$.015\\
w/o $\mathcal{L}_{\text{div}}$ \& adapters  &   .201$\pm$.019 &   .282$\pm$.035  &    .232$\pm$.020   \\
frozen adapters  &   .305$\pm$.055 &   .394$\pm$.008&    .320$\pm$.017     \\
rand.~init.~adapters  & .148$\pm$.070  &  .371$\pm$.012  &    .283$\pm$.026  \\
 \midrule
   \model\   & \textbf{.327}$\pm$.051  &  \textbf{.405}$\pm$.018 &    \textbf{.325}$\pm$.022  \\  
\bottomrule
\end{tabular}
\end{table}

\subsection{Ablation Study}
To investigate the impact of our designs, we study four variants of our model, as shown in Table~\ref{ablation}.

(1) \textbf{W/o $\mathcal{L}_{\text{div}}$}: %To further evaluate the importance of preserving modality-specific uniqueness, we conduct an ablation study where 
We remove the diversity loss \(\mathcal{L}_{\text{div}}\) from \model\ while retaining the modality adapters. This results in a pronounced drop in performance, indicating that the diversity loss is essential and effective for leveraging diverse and complementary information across modalities.
%, with an average MRR decrease of 13.3\% compared to \model. The findings indicate that removing $\mathcal{L}_{\text{div}}$ diminishes the model's ability to exploit complementary information across modalities, leading to a decline in predictive accuracy.

(2) \textbf{W/o $\mathcal{L}_{\text{div}}$ \& adapters}: 
We further remove the modality adapters in both meta-training and meta-testing.
%To further evaluate the contribution of our multimodal adapters, we conduct an ablation study where we remove all adapters during both meta-training and meta-testing. 
Instead, we use the same SVD-based fusion as applied to the traditional methods. This variant is equivalent to the MetaR baseline with SVD-based fusion. We observe a further decrease in performance, 
%This results in a substantial performance drop,with an average MRR decrease of 32.7\% compared to \model\, 
indicating that the simple fusion is prone to misalignment and noise across modalities. Thus, our adapters play a crucial role in adapting multimodal information to few-shot relations.
%The findings highlight the necessity of learnable adapters, which enable dynamic and optimized modality fusion for few-shot relation learning.

(3) \textbf{Frozen adapters}: To assess if the adapters can be fine-tuned for the test relations, we conduct an ablation study where the adapters are frozen in meta-testing, initialized from meta-trained adapters without further optimization. This leads to a notable decrease in MRR scores, averaging 3.66\% compared to \model. 
Hence, fine-tuning adapters during meta-testing is beneficial for adapting to test relations, even with just a few examples, owing to their parameter-efficient design.

(4) \textbf{Rand.~init.~adapters}: To further assess the contribution of meta-training to the adapters, 
we randomly initialize their parameters instead of using meta-trained adapters while allowing them to be updated in meta-testing. This leads to a more significant performance drop, averaging 25.3\% in MRR compared to \model.
This suggests that meta-training alone can learn relatively robust modality adaptation. This is because multimodal content is associated with entities, and entities appearing in unseen test relations are also present in abundant meta-training data. However, fine-tuning the adapters during meta-testing still plays an important role in generalizing to unseen relations when meta-learned initializations are used, as discussed with the third variant.

%, reaffirming that the primary performance gain comes from meta-training, where a larger dataset allows for effective learning. In contrast, the few-shot nature of meta-testing provides limited data for meaningful adaptation, making the randomly initialized adapters less effective. This underscores the importance of leveraging meta-trained adapters to ensure stable and effective modality fusion.

\begin{table}[tbp]
\centering
\caption{Comparison of unimodal and SVD-based multimodal fusion with few-shot learners in the five-shot setting, reporting MRR.}
\label{Comparative} 
\small
%\addtolength{\tabcolsep}{-1pt}
% \fontsize{8}{8}\selectfont
%
\begin{tabular}{l|ccc} 
  \toprule
   &  WN9-IMG  &  FB-IMG &  FB-IMG (F) \\  
   \midrule
MetaR (Unimodal) &   .209$\pm$.020 &    .288$\pm$.011 &   .246$\pm$.011  \\
MetaR (SVD) &   .201$\pm$.019 &    .282$\pm$.035 &   .239$\pm$.020   \\
\midrule
FAAN (Unimodal)   &   .234$\pm$.035 &   .295$\pm$.071 &   .245$\pm$.017  \\
FAAN (SVD)   &   .245$\pm$.034 &   .335$\pm$.046 &   .268$\pm$.037   \\
\midrule
HiRe (Unimodal)   &   .252$\pm$.039 &   .325$\pm$.091 &   .262$\pm$.048  \\
HiRe (SVD)   &   .238$\pm$.049 &   .307$\pm$.086 &   .256$\pm$.019   \\
\midrule
CIAN (Unimodal)   &   .241$\pm$.047 &   .315$\pm$.071 &   .283$\pm$.031  \\
CIAN (SVD)   &   .231$\pm$.049 &   .294$\pm$.106 &   .295$\pm$.038  \\
\midrule
MMSN (Unimodal)   &   .221$\pm$.041 &   .318$\pm$.056 &   .264$\pm$.045  \\
MMSN (SVD)   &   .231$\pm$.049 &   .294$\pm$.106 &   .255$\pm$.041   \\
\midrule
\model\ w/o $\mathcal{L}_{\text{div}}$   &  .275$\pm$.037 &    .375$\pm$.023 &   .307$\pm$.028 \\  
\model\   & \textbf{.327}$\pm$.051  &  \textbf{.405}$\pm$.018  &   \textbf{.325}$\pm$.022 \\  
\bottomrule
\end{tabular}%
\end{table}

\begin{table}[hbt!]
\centering
\caption{Modality combination analysis on different modalities for \model\ in five-shot setting, reporting MRR.}
\label{modality}
\small
%\setlength{\tabcolsep}{2pt}
% \fontsize{8}{8}\selectfont
%
\begin{tabular}{l|ccc}
\toprule
Modalities & WN9-IMG & FB-IMG & FB-IMG (F) \\
\midrule
Struct + Text & .253$\pm$.041 & .360$\pm$.016 & .317$\pm$.012 \\
Struct + Image & .238$\pm$.057 & .349$\pm$.014 & .272$\pm$.009 \\
Struct + Text + Image & \textbf{.327}$\pm$.051 & \textbf{.405}$\pm$.018 & \textbf{.325}$\pm$.022 \\
\bottomrule
\end{tabular}%
\end{table}

\subsection{Comparative Analysis for Fusion}

We further evaluate the effectiveness of our fusion module on multimodal data by first comparing it with unimodal and SVD-based fusion approaches, followed by examining the effect of combining different modalities.

\stitle{Fusion vs.~unimodal performance.}  
Table~\ref{Comparative} compares various FSRL methods under two settings: (i) \textit{Unimodal}, using only structural information, and (ii) \textit{SVD-based fusion}, which integrates structural, textual, and visual modalities. In general, the non-parametric nature of SVD leads to suboptimal fusion, as shown by the performance drop in models like MetaR and HiRe. SVD projects modalities into a shared space without explicitly aligning their feature representations, making it vulnerable to cross-modal misalignment and noise. As a result, modalities can interfere with each other rather than complement one another. A notable exception is FAAN, which improves under SVD fusion—likely due to its attention mechanism that filters out noisy neighboring entities in the KG. In contrast, MetaR and HiRe lack such a mechanism, making them more sensitive to multimodal noise.

Overall, \model\ significantly outperforms all baselines under both settings. Even “\model\ w/o $\mathcal{L}_{\text{div}}$” exceeds baseline performance, underscoring the effectiveness of our adapter-based fusion. By leveraging multimodal data through SVD-based fusion, FusionAdapter achieves an average MRR improvement of 10.9\% across all three datasets, clearly demonstrating the benefit of integrating diverse modalities.

\stitle{Effectiveness of modality combinations.}
In contrast, Table~\ref{modality} focuses on understanding the contribution of each modality by evaluating three configurations: (i) \textit{Struct + Text}, (ii) \textit{Struct + Image}, and (iii) \textit{Struct + Text + Image}.

The results consistently show that combining structure with text outperforms structure with image across all datasets, likely because textual data offers richer contextual cues and more nuanced relational semantics \cite{xie2017image}. Nevertheless, fusing all three modalities including structure, text, and image achieve the best performance. This comprehensive integration allows the model to capture a broader spectrum of relational knowledge, enhancing overall performance and highlighting the benefits of full multimodal fusion.

\subsection{Robustness analysis}\label{section:robustness}
\begin{figure}[t]
    \centering
    \includegraphics[width=0.8\linewidth]{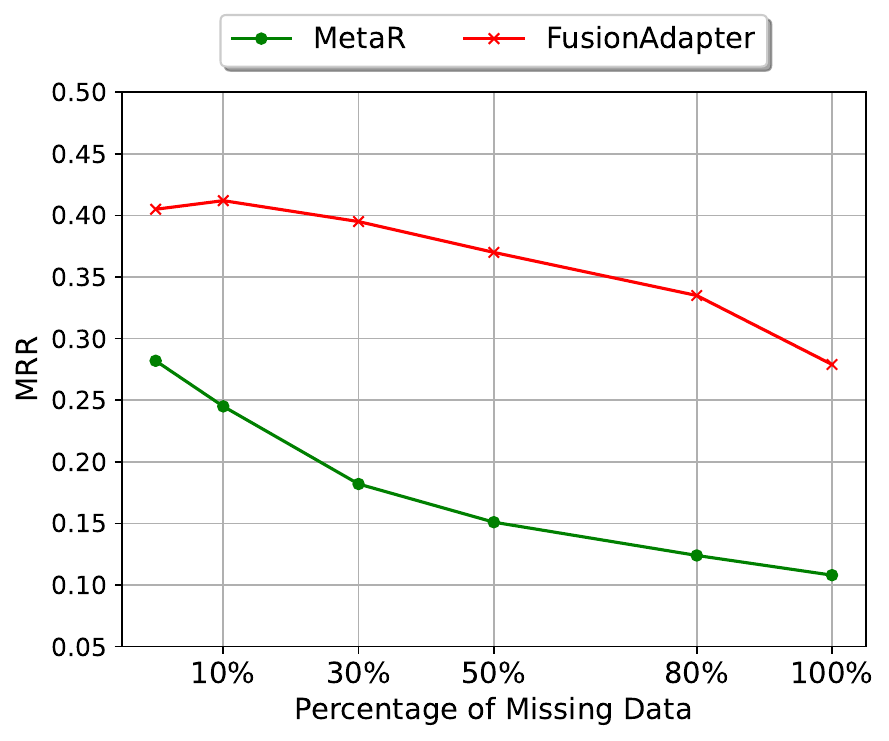}%
    \vspace{-1mm}
    \caption{Robustness analysis of the performance under different levels of missing modality data on FB-IMG, reporting MRR.}
    \label{fig:robustness}
\end{figure}

We evaluate the robustness of our approach under varying levels of missing modality data for FB-IMG dataset in Fig.~\ref{fig:robustness}.

Both models with \textit{SVD-based fusion} show a decline in performance as the proportion of missing modalities increases, highlighting the importance of multimodal information as complementary signals for relation prediction.

FusionAdapter exhibits a more gradual performance degradation compared to MetaR across all missing data levels. This indicates that (a) The modality-specific adapters in FusionAdapter effectively align partially missing or noisy modalities with structural features.(b) The diversity loss prevents partially informative modalities from collapsing into irrelevant or redundant representations. (c) While the meta-learning framework provides a baseline robustness in both models, FusionAdapter gains additional resilience through its multimodal fusion mechanisms.

\subsection{Efficiency Analysis}\label{section:efficiency}
We next analyze the efficiency of our approach in terms of parameter counts and runtime costs. %followed by measuring the actual runtime for the meta-train and meta-test stage. Results demonstrate the parameter efficient design of the adapter module.
\begin{table}[tbp] 
\centering
\caption{Parameter counts of our adapters and MetaR.}
%\vspace{-2mm}
\label{table:parametersize}
\small
%\addtolength{\tabcolsep}{-.5mm}
% \fontsize{8}{8}\selectfont
\begin{tabular}{c|ccc}
  \toprule
 % & \multicolumn{3}{c}{Number of parameters}  \\ \cmidrule{2-4}
 & WN9-IMG & FB-IMG & FB-IMG (F)   \\  \midrule
   MetaR  &  34,917,846  &  61,749,990 & 60,315,998     \\
   Our adapters    &   \phantom{0,}265,100  & \phantom{0,}265,100 & \phantom{0,}132,400  \\
 \midrule
  \% of MetaR    &   0.75 & 0.43 & 0.22     \\
\bottomrule
\end{tabular}
\end{table}

\begin{figure*}[hbt!]
    \centering
    \includegraphics[width=0.3\textwidth]{Figures/legend.pdf}
    
    \vspace{1mm}
    \begin{minipage}[t]{0.245\textwidth}
        \centering
        \includegraphics[width=\linewidth]{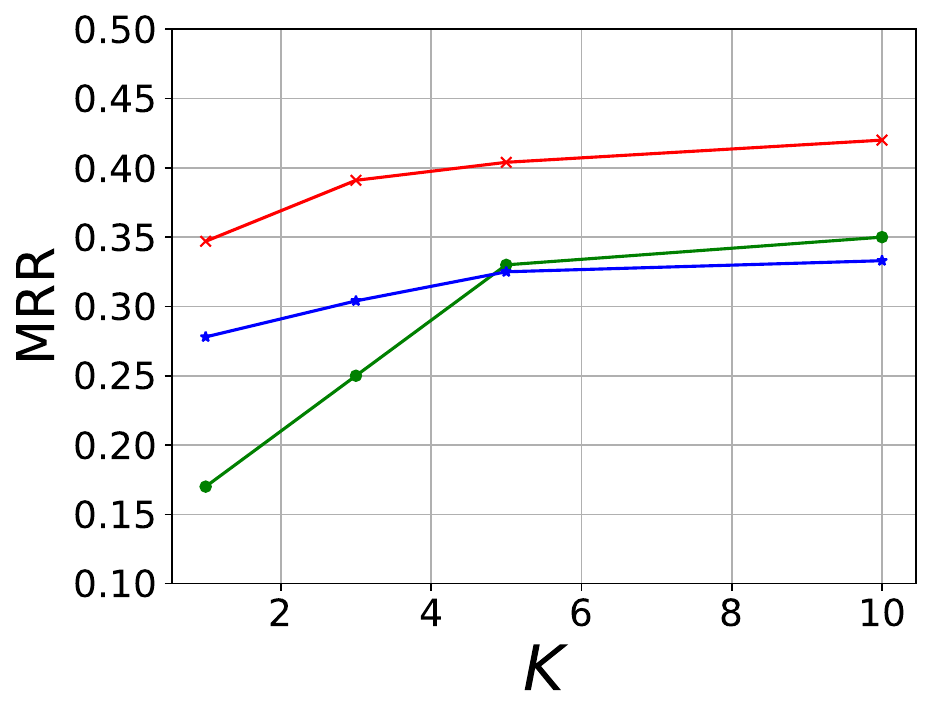}%
        \vspace{-1mm}
        \subcaption{Few-shot size}\label{fewshot}
    \end{minipage}
    \begin{minipage}[t]{0.245\textwidth}
        \centering
        \includegraphics[width=\linewidth]{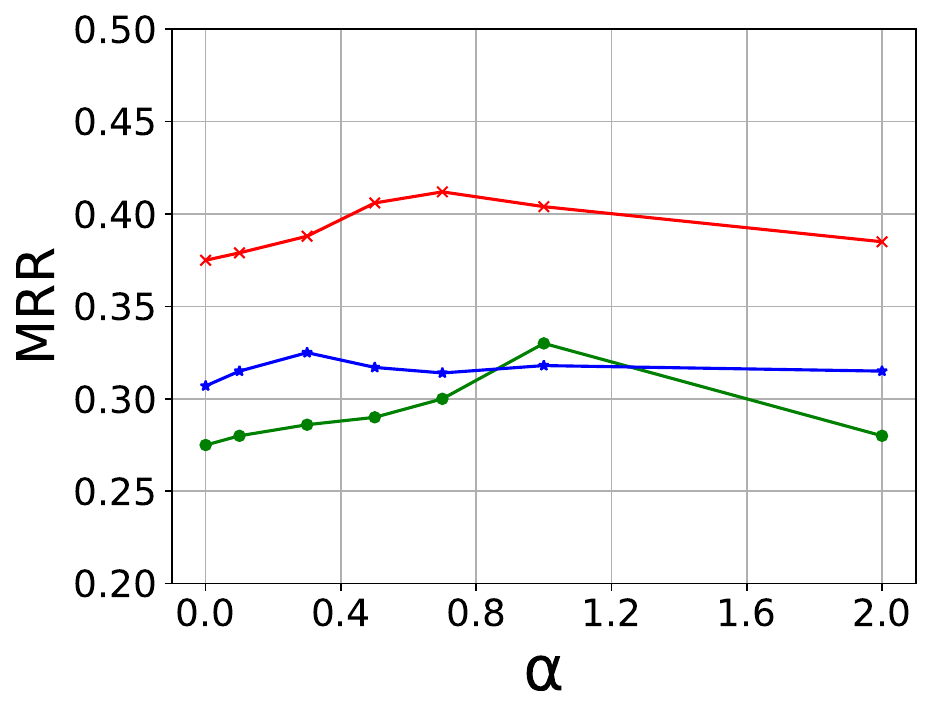}%
        \vspace{-1mm}
        \subcaption{Diversity coefficient}\label{adaptercoeffcient}
    \end{minipage}
    \begin{minipage}[t]{0.245\textwidth}
        \centering
        \includegraphics[width=\linewidth]{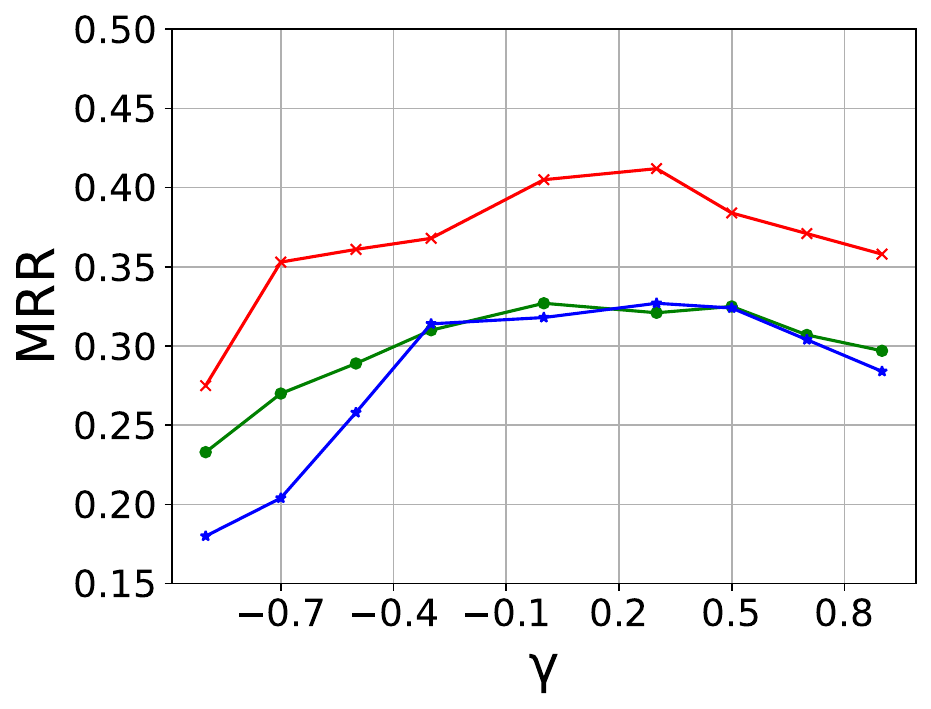}%
        \vspace{-1mm}
        \subcaption{Margin threshold}\label{margin_threshold}
    \end{minipage}  
    \begin{minipage}[t]{0.245\textwidth}
        \centering
        \includegraphics[width=\linewidth]{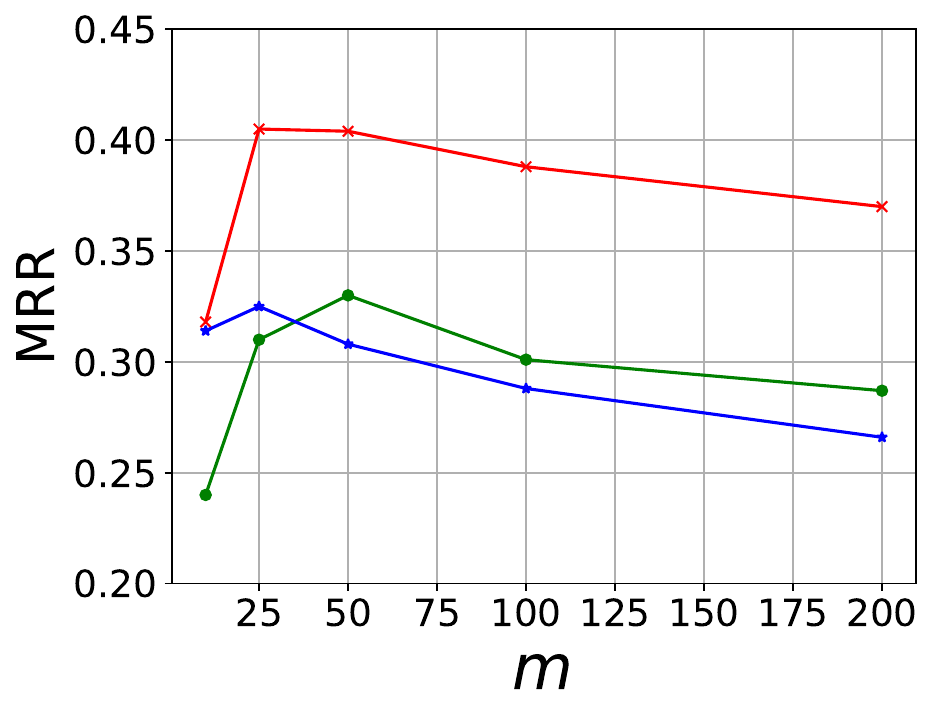}%
        \vspace{-1mm}
        \subcaption{Adapter neurons}\label{neurons}
    \end{minipage}
    \vspace{-1mm}
    \caption{Sensitivity analysis for the number of shots and hyperparameters.}
    \label{fig:1-2}
\end{figure*}

\stitle{Parameter efficiency.} A key advantage of our adapters is their parameter efficiency. As shown in Table~\ref{table:parametersize}, they add only a small number of parameters relative to MetaR. This parameter-efficient design reduces the risk of overfitting to the few-shot examples, enabling more robust generalization to unseen relations. Furthermore, compared to MetaR, \model\ introduces additional parameters only through the adapters, resulting in minimal overhead.

\stitle{Runtime efficiency.} As reported in Table~\ref{table:runtime}, \model\ generally incurs a lower or comparable meta-training time compared to other few-shot learners.  Despite the additional parameters introduced by the adapters, \model\ requires less meta-training time than HiRe and FAAN, which have more complex architectures. Compared to MetaR, which serves as our backbone architecture, our training time remains comparable.
%This demonstrates that the trade-off between additional parameters and runtime efficiency in \model\ is minimal, allowing it to outperform more resource-intensive models without a significant increase in training time.

\begin{table}[hbt!]
\centering
\caption{Comparison of meta-training and meta-testing runtimes (in seconds) for few-shot learners.}\label{table:runtime}
%\vspace{-2mm}
%\renewcommand{\arraystretch}{1.1}
%\addtolength{\tabcolsep}{-0.8mm}
\small
% \fontsize{8}{8}\selectfont
{\begin{tabular}{c|c|ccc}
  \toprule
Stage & Model & WN9-IMG & FB-IMG & FB-IMG (F)  \\  \midrule
   & FAAN  &   14,793 & 49,327  & 34,161   \\
    & HiRe  &   13,025 & 41,947  & 31,054  \\
   Meta-train& CIAN  & \phantom{0}8,721   &39,450 & 25,704  \\
    (total)  & MMSN  & \phantom{0}9,513   &   42,503& 34,797   \\
& MetaR  &  \phantom{0}7,081 & 30,665 & 25,462  \\
  & \model\    &   10,083 & 37,826  & 29,135 \\ \midrule
  & FAAN  &   0.036 & 0.09  & 0.11 \\
 & HiRe  &   0.025 & 0.06 & 0.05   \\
      Meta-test& CIAN  &  0.023  &  0.06 & 0.05   \\
    (per instance)& MMSN  & 0.026   &  0.07 & 0.07   \\
 & MetaR  &   0.017 & 0.03 & 0.05   \\
 & \model\    &   0.039 & 0.12  & 0.09 \\
\bottomrule
\end{tabular}}%
%}
\end{table}

Meanwhile, during meta-testing, \model\ requires more runtime than the baselines due to the additional fine-tuning of the adapters. However, the runtime remains within a comparable order of magnitude, with only a marginal absolute difference---especially when compared to the significantly longer meta-training stage.

\subsection{Sensitivity Analysis}\label{sec:expt:sensitivity}
We examine the impact of various settings and hyperparameters across three datasets, as illustrated in Fig.~\ref{fig:1-2}. In each figure, we vary one while keeping others fixed. The $x$-axis shows the range of values, while the $y$-axis shows the MRR scores.

\stitle{Few-shot size $K$.} This denotes the number of triples in the support set of each relation. As shown in Fig.~\ref{fig:1-2}(a), when $K$ increases, performance improves as expected, since more examples become available for fine-tuning during meta-testing.

\stitle{Diversity coefficient $\alpha$.}
As defined in Eqs.~\eqref{eqn:meta-train} and \eqref{eqn:meta-test}, \model\ introduces the hyperparameter $\alpha$ to control the weight of the diversity loss. Higher values of $\alpha$ give more importance to preserving modality-specific information. As shown in Fig.~\ref{fig:1-2}(b), the best performance is obtained when $\alpha$ is around 1 for WN9-IMG and FB-IMG, and $\alpha = 0.3$ for FB-IMG (F).

% Generally, when $\alpha<0.5$ or $\alpha>1$, performance declines significantly, as too little or too much diversity can harm the fusion mechanism. 

\stitle{Margin threshold $\gamma$.}
As defined in Eq.~\eqref{eqn:diversityloss}, the hyperparameter $\gamma$ sets the margin in the diversity loss, determining the maximum allowed similarity between modality-specific and shared representations before incurring a penalty. As illustrated in Fig.~\ref{fig:1-2}(c), the performance is optimal when $\gamma$ is around zero, which balances partial alignment and diversity of different modalities.

% However, when $\gamma<0$, excessive penalization of similarities between different modalities can hinder the learning process and degrade model performance. Conversely, when $\gamma>0.3$, it reduces the penalty for similarity between different modalities, allowing them to become overly similar, undermining the model's ability to capture distinct modality-specific features.}

\stitle{Adapter neurons $m$.} We analyze how the number of neurons $m$ in the hidden layer of the adapter network affects model performance. As shown in Fig.~\ref{fig:1-2}(d), the optimal $m$ is around 25--50 for both datasets. Increasing $m$ further reduces performance due to the significantly higher number of parameters, which may lead to overfitting and reduced generalization. %. More neurons capture complex patterns during meta-training, but excessive neurons can lead to overfitting, particularly during meta-testing, diminishing generalization. 

\section{Conclusion}
In this paper, we introduced \model, a multimodal fusion approach that aligns multiple modalities in knowledge graphs while preserving the unique characteristics of each modality. \model\ is equipped with a lightweight adapter module that adapts to different modalities in a parameter-efficient manner, enabling effective generalization to unseen relations in few-shot relation learning tasks. Moreover, we proposed a diversity loss to ensure that diverse and complementary features can be extracted from different modalities. Finally, we conducted comprehensive experiments on two benchmark datasets to demonstrate the superior performance of \model.

In future work, we plan to extend our method beyond the MetaR framework. While the current integration is limited to MetaR, the proposed method is model-agnostic and can be readily extended to other few-shot relation learning frameworks \cite{zhang2020few,sheng2020adaptive,niu2021relational,wu2023hierarchical}.

\appendix

\section*{Appendices}

\begin{table*}[t]
\small
\caption{Tuned hyperparameter settings based on validation data.}
%\vspace{-2mm}
\label{hyperparameters}
\begin{center}
\begin{tabular}{c| c c c c c} 
 \toprule
  & Hyperparameters & Range of values & WN9-IMG & FB-IMG & FB-IMG (F)\\
 \midrule
 TransE & norm & 1,2 & 1 & 2 &2  \\ 
 
DistMult & norm & 1,2  & 1 & 2 & 2\\

 ComplEx & norm & 1,2 & 2 & 2  &1 \\

 OTKGE & norm & 1,2 & 1 & 2 & 1 \\ 

TransAE & h,L & $1\sim10$  & 3 & 9 & 7 \\

 AdaMF-MAT & $\lambda$ &  $0.1,0.5, 1.0$ & 0.1 & 0.5 & 0.5 \\

 MetaR & beta & 1,3,5,10  & 3 & 3 & 5 \\

 FAAN & dropout\_input & 0.1,0.3,0.5,0.8 & 0.5 & 0.3 & 0.1\\

 HIRE  & beta & 1,3,5,10 & 5 & 5 & 5\\ 

 \model\ & $\alpha$, $m$ & $0.1 \sim 2.0$, $25 \sim 200 $ & $\alpha$ = 1.0, $m$ = 50 & $\alpha$ = 1.0, $m$ = 25 & $\alpha$ = 0.3, $m$ = 25  \\ 
 \bottomrule
\end{tabular}
\end{center}
\end{table*}

\section{codebases of baseline methods}\label{app:codebases}

For baseline methods, we use implementations from the following codebases.
\begin{itemize}[leftmargin=*]
    \item For traditional supervised models, including TransE, DistMult and ComplEx, we use their implementations from OpenKE\footnote{https://github.com/thunlp/OpenKE}.
    \item For multimodal supervised models, we adopt OTKGE\footnote{https://github.com/Lion-ZS/OTKGE}, TransAE\footnote{https://github.com/ksolaiman/TransAE}, AdaMF-MAT\footnote{https://github.com/zjukg/AdaMF-MAT} and NATIVE\footnote{https://github.com/zjukg/NATIVE}.
    \item For FSRL methods, we adopt  MetaR\footnote{https://github.com/AnselCmy/MetaR}, FAAN\footnote{https://github.com/JiaweiSheng/FAAN}, HiRE\footnote{https://github.com/alexhw15/HiRe}, CIAN\footnote{https://github.com/cjlyl/FKGC-CIAN} and MMSN\footnote{https://github.com/YuyangWei/MMSN}.
\end{itemize}

\section{Implementation Details}\label{Implementation_details}
For the WN9-IMG, FB-IMG and FB-IMG (F) datasets, we utilize the visual and textual pre-trained embeddings provided in \cite{cao2022otkge}. For structural and relational embeddings (when necessary), we employ the TransE-PyTorch implementation by Mklimasz\footnote{https://github.com/mklimasz/TransE-PyTorch}. Throughout our experiments, the embedding dimension is fixed at 100 for both datasets, and results are reported on the candidate set after removing relations with fewer than 20 candidates.

\stitle{Settings of \model.} We train \model\ for 100,000 epochs, evaluating on the validation set every 1,000 epochs and employing early stopping with a patience of 30. Mini-batch gradient descent is applied with a batch size of 1024 for both datasets. We use the Adam optimizer \cite{kingma2014adam} with an initial learning rate of 0.001 for the main model and 0.0001 for the adapters. The gradient update intensity is fixed at 5. In each query set, the number of positive and negative triplets is set to 3. The diversity coefficient, $\alpha$, and the number of hidden neurons, $m$, are tuned on the validation set (see Table~\ref{hyperparameters} for details).

\stitle{Settings of Baselines.} Key hyperparameters for all other baselines are tuned on the validation set, while the remaining settings follow their respective original papers. Refer to Table~\ref{hyperparameters} for further details.

\bibliographystyle{ACM-Reference-Format}
\bibliography{references}
\end{document}